\documentclass{article}

\usepackage{arxiv}

\usepackage[utf8]{inputenc} 
\usepackage[T1]{fontenc}    
\usepackage{hyperref}       
\usepackage{url}            
\usepackage{booktabs}       
\usepackage{amsfonts}       
\usepackage{nicefrac}       
\usepackage{microtype}      
\usepackage{lipsum}
\usepackage{graphicx}
\usepackage{subfig}

\hypersetup{
    colorlinks=true,
    linkcolor=blue,
    filecolor=blue,      
    urlcolor=blue,
    citecolor=blue
}

\usepackage{algorithm}
\usepackage[noend]{algpseudocode}
\usepackage{caption}

\algnewcommand\algorithmicforeach{\textbf{for each}}
\algdef{S}[FOR]{ForEach}[1]{\algorithmicforeach\ #1\ \algorithmicdo}

\title{Closed-Form Expressions for Global and Local Interpretation of Tsetlin Machines with Applications to Explaining High-Dimensional Data}

\author{ Christian D.~Blakely\thanks{Author's status: {\it Head of Artificial Intelligence and Real-Time Analytics}, PwC Switzerland. Any statements, ideas, and/or opinions expressed in this paper are strictly those of the author and do not necessarily reflect those of PwC Switzerland. The author can be contacted at: PwC Switzerland, Zurich, Switzerland 8004. E-mail: \texttt{christian.blakely@ch.pwc.com}.}\\
    Artificial Intelligence and Real-Time Analytics
	\\
	PwC Switzerland \\
	Zurich, Switzerland
	\And
	Ole-Christoffer~Granmo\thanks{Author's status: {\it Professor}. The author can be contacted at: Centre for Artificial Intelligence Research (\href{https://cair.uia.no}{https://cair.uia.no}), University of Agder, Grimstad, Norway.  E-mail: {\tt ole.granmo@uia.no}.}\\
	Centre for Artificial Intelligence Research (CAIR) \\
	University of Agder \\
	Grimstad, Norway
}

\hypersetup{
pdftitle={Closed-Form Expressions for Global and Local Interpretation of Tsetlin Machines with Applications to Explaining High-Dimensional Data},
pdfsubject={Interpretable AI},
pdfauthor={Christian D.~Blakely, Ole-Christoffer~Granmo},
pdfkeywords={Interpretable AI, Data Clustering, Dimension Reduction},
}

\date{July 2020}

\begin{document}

\maketitle

\begin{abstract}
Tsetlin Machines (TMs) capture patterns using conjunctive clauses in propositional logic, thus facilitating interpretation. However, recent TM-based approaches mainly rely on inspecting the full range of clauses \emph{individually}. Such inspection does not necessarily scale to complex prediction problems that require a large number of clauses. In this paper, we propose closed-form expressions for understanding why a TM model makes a specific prediction (local interpretability). Additionally, the expressions capture the most important features of the model overall (global interpretability). We further introduce expressions for measuring the importance of feature value ranges for continuous features. The expressions are formulated directly from the conjunctive clauses of the TM, making it possible to capture the role of features in real-time, also during the learning process as the model evolves.
Additionally, from the closed-form expressions, we derive a novel data clustering algorithm for visualizing high-dimensional data in three dimensions. Finally, we compare our proposed approach against SHAP and state-of-the-art interpretable machine learning techniques. For both classification and regression, our evaluation show correspondence with SHAP as well as competitive prediction accuracy in comparison with XGBoost, Explainable Boosting Machines, and Neural Additive Models.
\end{abstract} \hspace{10pt}

\section{Introduction}
Computational predictive modelling is becoming increasingly complicated in order to handle the deluge of high-dimensional data in machine learning and data science-driven industries. With rising complexity, not understanding why a model makes a particular prediction is becoming one of the most significant risks \cite{rudin2019stop}.  
To overcome this risk and to give insights into how a model is making predictions given the underlying data, several efforts over the past few years have provided methodologies for explaining so-called black-box models\footnote{We will understand black-box models as models which lack intrinsic interpretability features, such as ensemble approaches, neural networks, and random forests.}. The purpose is to offer performance enhancements over more simplistic, but transparent and interpretable models, such as linear models, logistic regression, and decision trees. Prominent efforts include SHAP \cite{lundberg2017unified}, LIME \cite{2016arXiv160204938T}, and modifications or enhancements to neural networks, such as in Neural Additive Models \cite{agarwal2020neural}. Typical approaches either create {\it post hoc} approximations of how models make a decision or build local surrogate models. As such, they require additional computation time and are not intrinsic to the data modelling or learning itself. In fact, in the recent paper \cite{2019arXiv191102508S}, it was demonstrated how easy it is to fool such approximations. 

Apart from increased trust, attaining interpretability in machine learning is essential for several reasons. For example, it can be useful when forging an analytical driver that provides insight into how a model may be improved, from both a feature standpoint and also a validation standpoint. It can also support understanding the model learning process and how the underlying data is supporting the prediction process \cite{interpretabilitySteps2019}. Additionally, interpretability can be used when reducing the dimensionality of the input features.

This paper introduces an alternative methodology for high-accuracy interpretable predictive modelling. Our goal is to combine competitive accuracy with closed-form expressions for both global and local interpretability, without requiring any additional local linear explanation layers or inference from any other surrogate models. That is, we intend to provide accessibility to feature strength insights that are intrinsic to the model, at any point during a learning phase. To this end, the methodology we propose enhances the intrinsic interpretability of the recently introduced Tsetlin Machines (TMs) \cite{granmo2018tsetlin}, which have obtained competitive results in terms of accuracy \cite{berge2019text,abeyrathna2020regression,granmo2019convolutional}, memory footprint \cite{granmo2019convolutional,wheeldon2020learning}, energy \cite{wheeldon2020learning}, and learning speed \cite{granmo2019convolutional,wheeldon2020learning} on diverse benchmarks (image classification, regression and natural language understanding).

TMs are a novel machine learning approach that combines elements of reinforcement learning, learning automata \cite{Narendra1989}, and game theory \cite{von2007theory} to obtain intricate pattern recognition systems. They use frequent pattern mining \cite{Haugland2014} and resource allocation principles \cite{Granmo2007d} to extract common patterns in the data, rather than relying on minimizing output error, which is prone to overfitting.  Unlike the intertwined nature of pattern representation in neural networks, a TM decomposes problems into self-contained patterns, expressed as conjunctive clauses in propositional logic (e.g., in the form \textbf{if} X \textbf{satisfies} condition A \textbf{and not} condition B \textbf{then} Y = 1). The clause outputs, in turn, are combined into a classification decision through summation and thresholding, akin to a logistic regression function, however, with binary weights and a unit step output function \cite{berge2019text}. Being based on disjunctive normal form (DNF), like Karnaugh maps, the TM can map an exponential number of input feature value combinations to an appropriate output \cite{phoulady2020weighted}.

\textbf{Recent progress on TMs.} Recent research reports several distinct TM properties. The TM can be used in convolution, providing competitive performance on MNIST, Fashion-MNIST, and Kuzushiji-MNIST, in comparison with CNNs, K-Nearest Neighbor, Support Vector Machines, Random Forests, Gradient Boosting, BinaryConnect, Logistic Circuits and ResNet \cite{granmo2019convolutional}. The TM has also achieved promising results in text classification using the conjunctive clauses to capture textual patterns \cite{berge2019text}. Further, by introducing real-valued clause weights, the number of clauses can be reduced by up to $50\times$ without loss of accuracy \cite{phoulady2020weighted}. Being based on logical inference, indexing the clauses on the features that falsify them increases inference- and learning speed by up to an order of magnitude \cite{gorji2020indexing}. The hyper-parameter search of a TM involves three hyper-parameters. Multi-granular clauses simplify the hyper-parameter search by eliminating the pattern specificity parameter \cite{gorji2019multigranular}. Recently, the Regression TM compared favorably with Regression Trees, Random Forest Regression, and Support Vector Regression \cite{abeyrathna2020regression,abeyrathna2019scheme}. In \cite{abeyrathna2020extending}, stochastic searching on the line (SSL) automata \cite{oommen1997stochastic} learn integer clause weights, performing competitively against Random Forest, Gradient Boosting, Explainable Boosting Machines, as well as the standard TM. Finally, \cite{shafik2020explainability} shows that TMs can be fault-tolerant, completely masking stuck-at faults.

\textbf{Paper Contributions.} Interpretability in all of the above TM-based approaches relies on inspecting the full range of clauses \emph{individually}. Such inspection does not necessarily scale to complex pattern recognition problems that require a large number of clauses, e.g., in the thousands. A principled interface for accessing different types of interpretability at various scales is thus currently missing.  In this paper, we introduce closed-form expressions for local and global TM interpretability. We formulate these expressions at both an overall feature importance level and a feature range level, namely, which ranges of the data yield the most influence in making predictions. Secondly, we evaluate performance on several industry-standard benchmark data sets, contrasting against other interpretable machine learning methodologies. We further compare the interpretability results in terms of global feature importance. Finally, we provide an application of our interpretability approach by deriving a clustering method from the local feature importance of the model. Such clustering could be used for visualizing high-dimensional tabular data in two or three dimensions, for example.

\section{Tsetlin Machine Basics}\label{sec:basics}

\subsection{Classification}

\begin{figure}[b!!]
\centering
\includegraphics[width=0.6\textwidth]{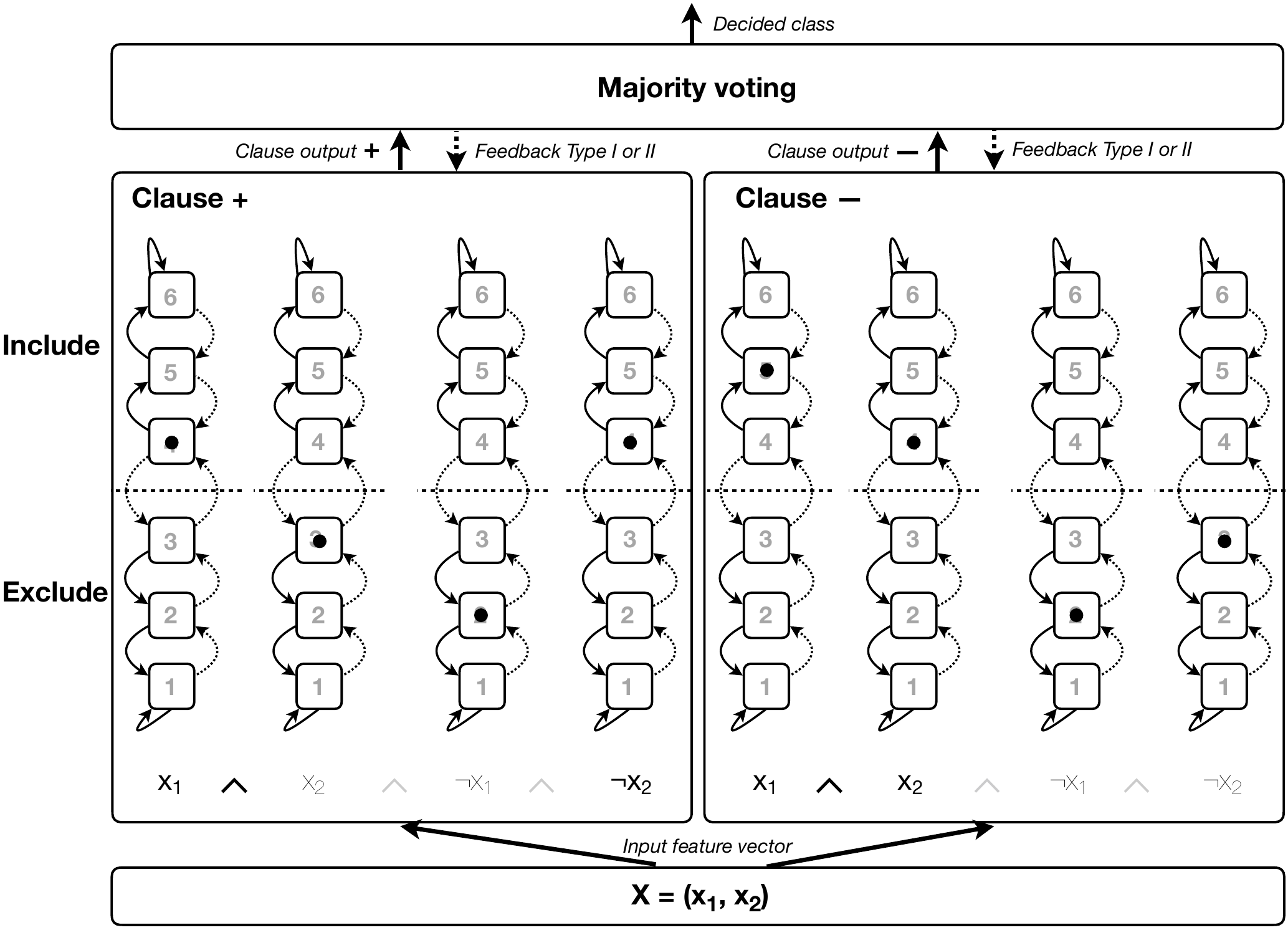}
\caption{Two TA teams, each producing a conjunctive clause. Classification is based on majority voting. }\label{figure:tm_architecture}
\end{figure}

A TM takes a vector $X=(x_1,\ldots,x_o)$ of Boolean features as input (Figure \ref{figure:tm_architecture}), to be classified into one of two classes, $y=0$ or $y=1$. Together with their negated counterparts, $\bar{x}_k = \lnot x_k = 1-x_k$, the features form a literal set $L = \{x_1,\ldots,x_o,\bar{x}_1,\ldots,\bar{x}_o\}$. In the following, we will use the notation $I_j$ to refer to the indexes of the non-negated features in $L_j$ and  $\bar{I}_j$ to refer to the indexes of the negated features.

A TM pattern is formulated as a conjunctive clause $C_j$, formed by ANDing a subset $L_j \subseteq L$ of the literal set:
\begin{equation}
C_j (X)=\bigwedge_{l_k \in L_j} l_k = \prod_{l_k \in L_j} l_k.
\end{equation}
E.g., the clause $C_j(X) = x_1 \land x_2 = x_1 x_2$ consists of the literals $L_j = \{x_1, x_2\}$ and outputs $1$ iff $x_1 = x_2 = 1$.

The number of clauses employed is a user set parameter $m$. Half of the clauses are assigned positive polarity. The other half is assigned negative polarity. In this paper, we will indicate clauses with positive polarity with odd indices and negative polarity with even indices \footnote{Any systematic division of clauses can be used as long as the cardinality of the positive and negative polarity sets are equal.}. The clause outputs, in turn, are combined into a classification decision through summation and thresholding using the unit step function $u(v) = 1 ~\mathbf{if}~ v \ge 0 ~\mathbf{else}~ 0$:
\begin{equation}
\hat{y} = u\left(\sum_{j=1, 3, \ldots, m-1} C_j(X) - \sum_{j=2,4,\ldots, m} C_j(X)\right).
\end{equation}
Namely, classification is performed based on a majority vote, with the positive clauses voting for $y=1$ and the negative for $y=0$. The classifier
$\hat{y} = u\left(x_1 \bar{x}_2 + \bar{x}_1 x_2 - x_1 x_2 - \bar{x}_1 \bar{x}_2\right)$, e.g., captures the XOR-relation.

\subsection{Learning}

A clause $C_j(X)$ is composed by a team of Tsetlin Automata \cite{Tsetlin1961}, each Tsetlin Automaton deciding to \emph{Include} or \emph{Exclude} a specific literal $l_k$ in the clause (see Figure \ref{figure:tm_architecture}). Learning which literals to include is based on reinforcement: Type I feedback produces frequent patterns, while Type II feedback increases the discrimination power of the patterns.

A TM learns on-line, processing one training example $(X, y)$ at a time.

\textbf{Type I feedback} is given stochastically to clauses with positive polarity when $y=1$ and to clauses with negative polarity when $y=0$. An afflicted clause, in turn, reinforces each of its Tsetlin Automata based on:
\begin{enumerate}
    \item The clause output $C_j(X)$;
    \item The action of the targeted Tsetlin Automaton -- \emph{Include} or \emph{Exclude}; and
    \item The value of the literal $l_k$ assigned to the automaton.
\end{enumerate}
Two rules govern Type I feedback:
\begin{itemize}
\item \emph{Include} is rewarded and \emph{Exclude} is penalized with probability $\frac{s-1}{s}~\mathbf{if}~C_j(X)=1~\mathbf{and}~l_k=1$. This reinforcement is strong (triggered with high probability) and makes the clause remember and refine the pattern it recognizes in $X$. 
\item \emph{Include} is penalized and \emph{Exclude} is rewarded with probability $\frac{1}{s}~\mathbf{if}~C_j(X)=0~\mathbf{or}~l_k=0$. This reinforcement is weak (triggered with low probability) and coarsens infrequent patterns, making them frequent.
\end{itemize}
Above, $s$ is a hyperparameter that controls the frequency of the patterns produced.

\textbf{Type II feedback} is given stochastically to clauses with positive polarity when $y=0$ and to clauses with negative polarity when $y=1$. It penalizes \emph{Exclude} with probability $1$ $\mathbf{if}~C_j(X)=1~\mathbf{and}~l_k=0$. This feedback is strong and produces candidate literals for discriminating between $y=0$ and $y=1$.

\begin{figure}[b!!]
\centering
\includegraphics[width=0.7\textwidth]{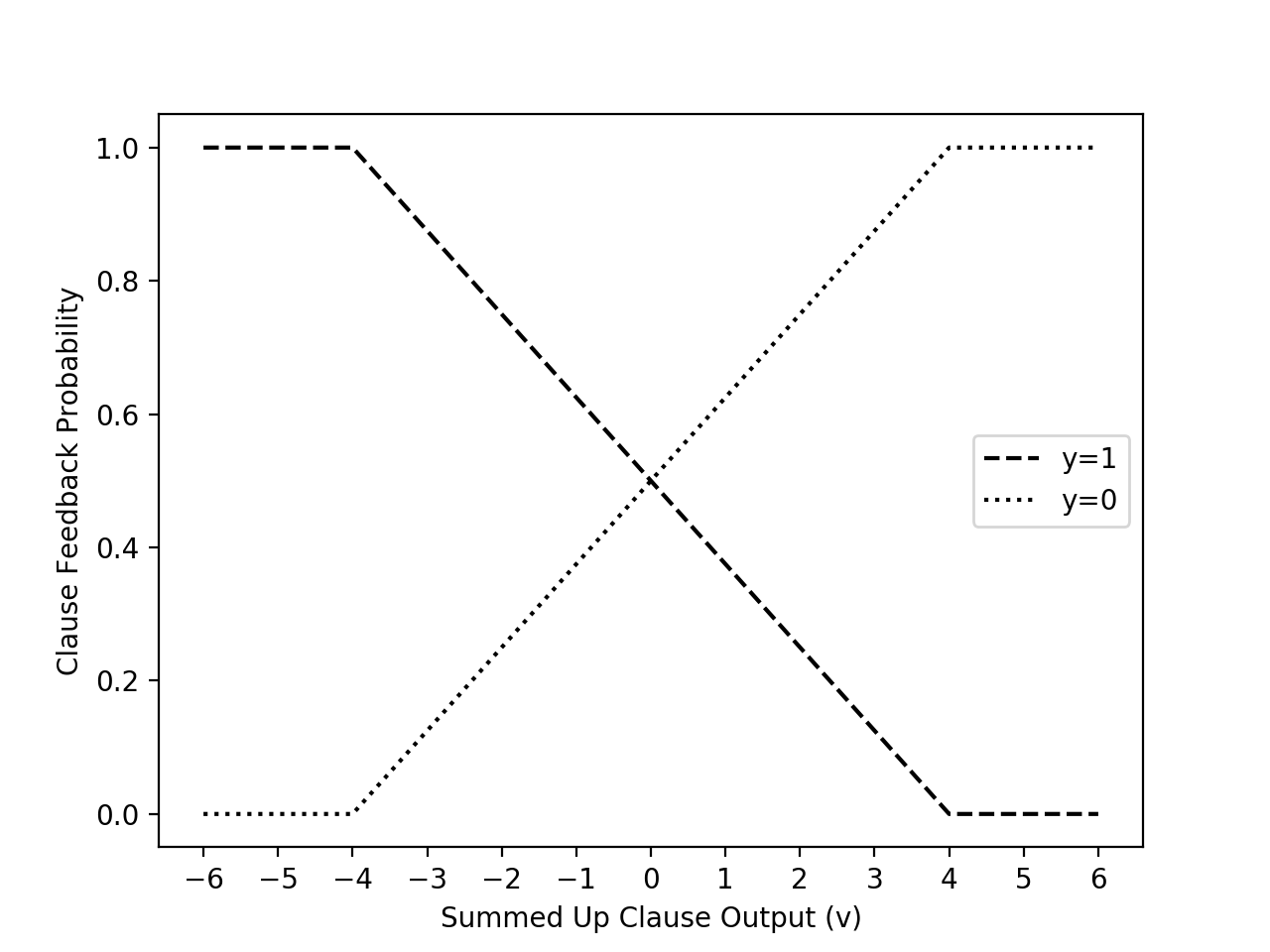}
\caption{Clause feedback probability for $T=4$.}\label{figure:clause_activation_probability}
\end{figure}

\textbf{Resource allocation dynamics} ensure that clauses distribute themselves across the frequent patterns, rather than missing some and overconcentrating on others. That is, for any input $X$, the probability of reinforcing a clause gradually drops to zero as the clause output sum
\begin{equation}
    v = \sum_{j=1, 3, \ldots, m-1} C_j(X) - \sum_{j=2,4,\ldots, m} C_j(X)
\end{equation}
approaches a user-set target $T$ for $y=1$ (and $-T$ for $y=0$). To exemplify, Figure \ref{figure:clause_activation_probability} plots the probability of reinforcing a clause when $T=4$ for different clause output sums $v$, per class $y$. If a clause is not reinforced, it does not give feedback to its Tsetlin Automata, and these are thus left unchanged.  In the extreme, when the voting sum $v$ equals or exceeds the target $T$ (the TM has successfully recognized the input $X$, no clauses are reinforced. Accordingly, they are free to learn new patterns, naturally balancing the pattern representation resources. See \cite{granmo2018tsetlin} for further details. 

\section{Tsetlin Machine Interpretability} 

In this section, we first introduce the current notion of TM interpretability, before we propose novel expressions for understanding a TM model from the viewpoint of the encoded input features and their prediction strength. Because the TM represents patterns as self-contained conjunctive clauses in propositional logic, the method naturally leads to straightforward interpretability.  However, as is the consensus in the machine learning literature, a model is fully interpretable only if it is possible to understand how the underlying data features impact the predictions of the model, both from a global perspective (entire model) and at the individual sample level. For example, a model that predicts housing prices in California should give much emphasis on specific latitude/longitude coordinates, as well as the age of the neighbouring houses (and perhaps other factors such as proximity at the beach). For individual sample data points, the model interpretation should show which features (such as house age, proximity to the beach) had the most positive (or negative) impact in expression the prediction. Before we present how closed-form expressions for both global and local interpretability can be derived intrinsically from any TM model, we first define some basic notation on the encoding of the features to a Boolean representation and feature inclusion sets, a critical first step in TM modeling.  

\subsection{Boolean Representation}

We first assume that the input features to any TM can be booleanized into a bit encoded representation. Each bit represents a Boolean variable that is either False or True ($0$ or $1$). To map continuous values to a Boolean representation, we use the encoding scheme proposed in \cite{abeyrathna2019scheme}, adapted as follows. Let $I_u = [a_u, b_u]$ be some interval on the real line representing a possible range of continuous values of feature $f_u$. Consider $l > 0$ unique values $u_1, \ldots, u_l \in I_u$. We encode any $v \in I_u$ into an $l$-bit representation $[\{0,1\}]^l$ where the $i$th bit is given as $1 \; \mbox{if} \; v \geq u_i$, and $0$ otherwise. We will denote $v \in I_u$ of a feature $f_u$ as $X_{f_u} := [x_1, \ldots, x_l]$ where any $x_i := v \geq u_i$. For example, suppose $I_u =[0,10]$ with $u_1 = 1, u_2 = 2, \ldots, u_{10} = 10$. Then $v = 5.5$ is encoded as $[1,1,1,1,1,0,0,0,0,0]$, while any $v <= 0$ becomes $[0,0,0,0,0,0,0,0,0,0]$, and any $v >= 10$ becomes $[1,1,1,1,1,1,1,1,1,1]$. 

The choice of $l > 0$ and the values $u_1, \ldots, u_l \in I_u$ should typically be chosen according to the properties of the empirical distribution of the underlying feature. If the values of feature $f_u$ are relatively uniform across $I_u$, then a uniform choice of $u_1, \ldots, u_l \in I_u$ would seem appropriate. Otherwise the grid should be chosen to accompany the underlying density of the data, with a finer grid choice near higher densities.

\subsection{Global Interpretability} 

Global interpretability is interested in understanding the most salient (important) features of the model, namely to what degree and strength a certain feature impacts predictions overall. We now introduce multiple output variables $y^i \in \{y
^1, y^2, \ldots, y^n\}$, $y^i = \{0,1\}$ and the upper index $i$, which refers to a particular output variable. For simplicity in exposition, we assume that the corresponding literal index sets $I^i_j, \bar{I}^i_j$, for each output variable, $i = 1, \ldots, n$, and clause, $j = 1, \ldots, m$, have been found under some performance criteria of the learning procedure, described in Section \ref{sec:basics}.  With these sets fixed, we can now assemble the closed form expressions. 

\textbf{Global Feature Strength.} For global feature strength we make direct use of the indexed inclusion sets $I^i_j, \bar{I}^i_j$, which are governed by the actions of the Tsetlin Automata. Specifically, we compute positive (negative) feature strength for a given output variable $y^i$ and $k$th bit of feature $f_u$ as follows:
\begin{equation}\label{global1}
g[k] \leftarrow \frac{1}{m} \sum_{j \in \{1,3,\ldots,m-1\}} \{1 \mbox{ if } k \in I^i_j\}, \; \; \; \;  \bar{g}[k] \leftarrow \frac{1}{m} \sum_{j \in \{1,3,\ldots,m-1\}} \{1 \mbox{ if } k \in \bar{I}^i_j\}.
\end{equation}
In other words, for any given feature index $k \in [1,\ldots,o]$, the frequency of its inclusion across all clauses, restricted to positive polarity, governs its global importance score $g[k]$. This score thus reflects how often the feature is part of a pattern that is important to making a certain class prediction $y^i$. Notice that we are only interested in indices pertaining to the positive polarity clauses $C^i_j(X)$ since these are the clauses that contain references to features that are beneficial for predicting the $i$th class index. 

The $g[k]$ and $\bar{g}[k]$ scores give positive and negative importance at the bit level for each feature. To get the total strength for a feature $f_u$ itself, we simply aggregate over each bit $k$ based on Eqn. \ref{global1}, restricted to the given feature $f_u$:
\begin{equation}\label{global2}
\phi(f_u) \leftarrow \sum_{k | x_k \in X_{f_u}} g[k], \; \; \; \; \bar{\phi}(f_u) \leftarrow \sum_{k | x_k \in X_{f_u}} \bar{g}[k].
\end{equation}
Above, $X_{f_u}$ should be understood as the bit encoded features representing feature $f_u$. The functions $\phi(\cdot)$ accordingly measure the general importance of a feature when it comes to predicting a given class label $y^i$. As we shall see empirically in the next section, this measure defines the most relevant features of a model.

\textit{Remark:} Due to the fact that the above expression can be trivially computed given any inclusion sets derived from the clauses, the feature importance can be observed in real-time during the online training procedure of TMs, to see how the feature importance evolves with new unseen date training samples. Such a feature is not available in typical batch machine learning methodologies utilizing frameworks such as SHAP. 

\textbf{Global Continuous Feature Range Strength.} One additional global feature importance representation that we can derive for continuous input feature ranges is measuring the importance of a feature in terms of ranges of values. To do this, we can make use of the inclusion of features and negated features to construct a mapping of the important ranges for a given class. Based on our Boolean representation of continuous features, we define the allowable range for $f_u$ as follows:
\begin{equation}\label{global3}
R(f_u) := [\mbox{min}_{k \in \bar{I}^i_j}  f_u(x_k), \mbox{max}_{k \in I^i_j}  f_u(x_k)].
\end{equation}

Notice that according to the literal $x_k$, $\mbox{max}_{k \in I^i_j}  f_u(x_k)$, gives the largest $u_k \in I_u$ such that $v \geq u_k$ was an influential feature for an input $v$. 
Similarly, $\mbox{min}_{k \in \bar{I}^i_j}  f_u(x_k)$ yields the smallest $u_k \in I_u$ such that $\lnot (v \geq u_k)$, or $v < u_k$, was a relevant feature for $y^i$. 

So now do we not only know the most salient features in making a certain prediction, but we know the range of the influential values as well for every feature $f_u$ given prediction class $y^i$. Again, these can all be computed in real-time during the learning of the model, giving full insight into the modeling procedure.

\subsection{Local Interpretability}

Local interpretability is interested in understanding to what degree features positively (or negatively) impact individual predictions. Due to the transparent nature of the clause structures, it is relatively straightforward to extract the driving features of a given input sample. If we assume again the bit representation $X = [x_1, x_2, \ldots, x_o ]$ is a concatenation of the bit representation of all features $f_{u_j}$, and we suppose $i$ is the predicted class index of $X$, then we define 
\begin{equation}\label{locall3}
l[k,X] \leftarrow \sum_{j \in \{1,3,\ldots,m-1\}} \{C_j^i(X) \, | \, x_k = 1,  k \in I^i_j \} 
\end{equation}
which give positive importance at the bit level for each feature. To get the total predictive impact for each feature we again aggregate over the individual bits of the feature:
\begin{equation}\label{locall4}
\phi(f_u, X) \leftarrow \sum_{k \, | \, x_k \in X_{f_u}} l[k,X], \; \; \; \; \bar{\phi}(f_u, X) \leftarrow \sum_{k \, | \, x_k \in X_{f_u}} \bar{l}[k,X].
\end{equation}   
In other words, for a given input, all the combined clauses act as a lookup table of important features, from which the strength of each feature in a prediction can be determined. 

\subsection{Data Dimension Reduction and Clustering}

In this section, we demonstrate a direct application of our closed-form expressions for global and local interpretability, namely, clustering and visualization of high dimensional data. To this end, we use both the global and local expressions, Eqn. \ref{global2} and Eqn. \ref{locall3}, respectively. In Algorithm \ref{alg:reduc}, we outline the basic procedure, and then go line by line on what each step is doing. The algorithm begins by assuming a collection of TM clauses have been learned on certain subset of the data, paired with its respective labels. In going forward, we will assume these labels are in the form of indices from $C$ different classes, and that we have $K$ different features of any data sample $x \in \mathbf{X}$.

\begin{algorithm}
\caption{Data dimension reduction and clustering}\label{alg:reduc}
\begin{algorithmic}[1]
\Procedure{Cluster}{$ \mathbf X$} 
\State \mbox{Input: }Class index $C$ 
\State \mbox{Input: } $m \leq K$ number of features
\State Generate $m$ centroids $\mathbf{o}_1,\ldots \mathbf{o}_m$ in $[0,1]^n$ 
\State Rank all features $u_i$ according to Eqn. \ref{global2}
\State Associate each $\mathbf{o}_i$ with $u_i$
\ForEach {$\mathbf{x} \in \mathbf X $}
\State Compute binarization of $\mathbf{x}$ giving $\mathbf{f}_u := (f_{u_1},\ldots,f_{u_n})$
\State Evaluate $\phi(f_{u_i}) \in \mathbf{f}_u$ for all $i$   
\ForEach {$\mathbf{o}_i$, $i < m$} 
\State If $u_i$ is largest contributing feature 
\State Then set $\mathbf{x}|_{u_i} := \mathbf{o}_i$
\EndFor
\EndFor
\ForEach {$\mathbf{o}_i$}
\ForEach {$\mathbf{x}|_{u_i}$} 
\State For $j$ largest factor $u_j$ of $\mathbf{x}$, shift $\mathbf{x}|_{u_i}$ in direction of $\mathbf{o}_j$ by $\frac{\| \mathbf{o}_i - \mathbf{o}_j \|_2}{j}$ 
\State Repeat for all observations over all features 
\EndFor
\EndFor
\State \textbf{return} $\mathbf{x}|_{u_1}$ for all $\mathbf{x} \in \mathbf X$
\EndProcedure
\end{algorithmic}
\end{algorithm}

The algorithm begins by allocating $m \leq K$ centroids, namely points of reference, in $[0,1]^n$. The dimension $n$ is typically taken as 2 or 3. In line 5 we are ranking all the $K$ features according to the global importance expression (Eqn. \ref{global2}). The centroid $\mathbf{o}_1$ is associated with the highest ranking feature, and the $\mathbf{o}_2$ centroid is associated with the second highest ranking features, and so forth. For each data sample, we compute the binarization encoding into bits, and evaluate the local interpretability expression (Eqn. \ref{locall4}). Taking the feature with the highest importance, say $u_i$, we then map the sample to the centroid $\mathbf{o}_i$, which we will refer to as $x|_{o_i}$. Taking the second feature with the highest importance, say $u_j$, we then translate $x|_{o_i}$ in the direction of $u_j$ by a scale of the Euclidian distance between the two centroids. We can repeat this procedure one or two more times, and then continue the same procedure over all samples in consideration. Eventually, all samples are attributed to a neighborhood of some localized independent centroid, and all samples are also in the vicinity of other samples which share very similar characteristics in regards to their local interpretability. So not only has the dimension of the data been reduced to 2 or 3 dimensions, and all the data has been clustered together in a visually meaningful and interpretable way, but one can understand various distribution properties of the data as well, for example, how biased it is towards a particular feature. 

\section{Empirical Evaluation and Applications}

In order to evaluate the capability of our expressions for global and local TM interpretability, we will now consider two types of data sets: 1) Data sets which have an intuitive set of features which are already quite interpretable without model insight and 2) Data sets which will be more data driven, with features that are not necessarily understood by humans. We will further compare our results with other interpretable approaches to gain insight into the properties of our closed-form expressions. We further demonstrate that the TM can compete accuracy-wise with black-box methods, while simultaneously remaining relatively transparent.

Throughout the section, we will be using the Integer Weighted Tsetlin Machine (IWTM) proposed in \cite{abeyrathna2020extending}, a recent extension of the classical TM summarized in Section \ref{sec:basics}.

\subsection{Comparison with SHAP}

We first compare our feature strength scores, as defined in Eqn. \ref{global2} and Eqn. \ref{global3}, with SHapley Additive exPlanations (SHAP) values, a popular methodology for explaining black-box models, considered as state-of-the-art. First published in 2017 by Lundberg and Lee \cite{lundberg2017unified}, the  approach attempts to "reverse-engineer" the output of any predictive algorithm by assigning each feature an importance value for a particular prediction. As it is first and foremost concerned with the local interpretability of a model, global interpretability of the SHAP approach can be viewed at the level of visualizing all the samples in the form of a SHAP value (impact on model output) by feature value (from low to high). 

We use the Wisconsin Breast Cancer data set to compare the two approaches. This data set contains 30 different features extracted from an image of a fine needle aspirate (FNA) of a breast mass. These features represent different physical characteristics like the concavity, texture, or symmetry computed for each cell with various descriptive statistics such as the mean, standard deviation and "worst" (sum of the three largest values). Each sample is labelled according to the diagnosis, benign or malign.

Before comparing the results on the most influential features, we compare the performance metrics of the IWTMs with XGBoost. Table \ref{compareShap} shows the comparison in terms of mean and variance on accuracy and error types where the models were constructed 10 times each, with a training set of 70 percent randomly chosen samples. We see that the IWTM clearly competes in performance with XGBoost, albeit with a much higher variance in performance.

\begin{table}
\centering
\caption{Performance comparison of the IWTM and XGBoost models on the breast cancer data set
}\label{compareShap}
\begin{tabular}{|| c | c | c | c | c ||}
\hline
Method & Accuracy & AUC & Type I Error & Type II Error \\
\hline
IWTM & 95.19 $\pm$ 0.52 & 98.10 $\pm$ 0.43 & 0.033 $\pm$ 0.021 & 0.053 $\pm$ 0.023 \\
XGBoost & 95.45 $\pm$ 0.15 & 98.02 $\pm$ 0.04 & 0.039 $\pm$ 0.01 & 0.054 $\pm$ 0.01 \\
\hline
\end{tabular}

\end{table}

\begin{figure}[ht]
\caption{SHAP value impact on model output against feature value plot}
\label{shapresults}
\centering
\includegraphics[width=.6\textwidth]{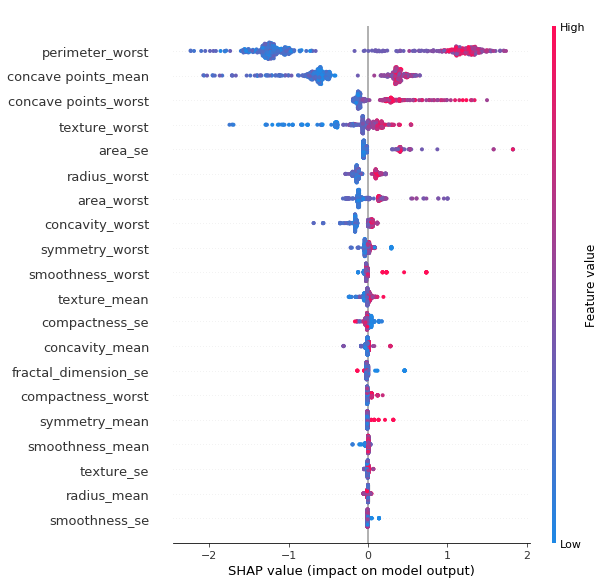}
\end{figure}

In the SHAP summary plot shown in Figure \ref{shapresults}, we observe that the most important features for the XGBoost model are  \verb|perimeter_worst|, \verb|concave points-mean|, \verb|concave points-worst|, \verb|texture_worst| and \verb|area_se| among others. 
Taking into account the feature value and impact on model output, the most impactful features for making a prediction tend to be \verb|perimeter_worst|,  \verb|concave points-worst|, \verb|area_se|, and \verb|area_worst|, whereas \verb|texture_worst| and \verb|concave points-mean| prove to be the most effective in their lower value range.

\begin{figure}[ht]
\caption{Global feature importance for predicting output of malign}
\label{globalMalign}
\centering
\includegraphics[width=.6\textwidth]{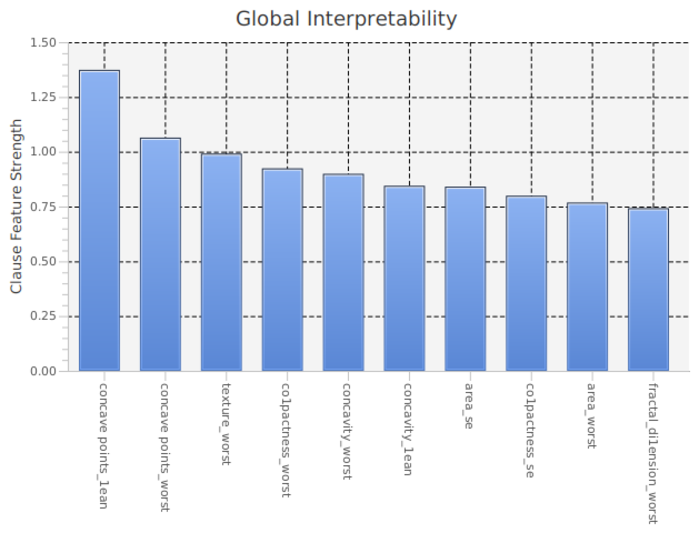}
\end{figure}

Figure \ref{globalMalign} shows the corresponding results for the TM, providing the strength of the features when predicting benign cells, calculated based on Eqn. \ref{global2}. We can see that \verb|concave points-mean|, \verb|concave points-worst|, \verb|texture-worst|, \verb|area_se|, and \verb|concavity_worst| seem to be the strongest contributors. Apart from feature \verb|perimeter_worst| being the most effective for SHAP, it can clearly be seen that both methods yield very similar global feature strengths. Nearly all of the top 10 features in both sets agree in relative strength.

\begin{figure}[ht]
\caption{Global feature importance for predicting output of benign}
\label{globalBenign}
\centering
\includegraphics[width=.6\textwidth]{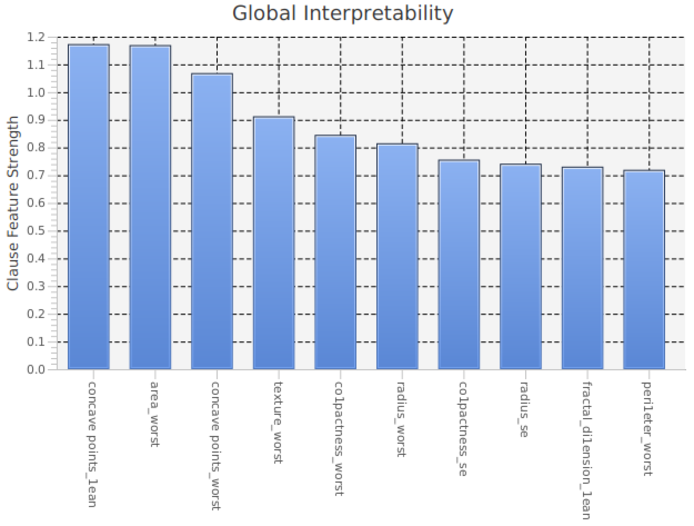}
\end{figure}

Similarly, the scores of the most relevant features for the benign class are shown in Figure \ref{globalBenign}. As seen, \verb|concave points-mean|, \verb|concave points-worst|, and \verb|area_worst| have a non-trivial impact.  

\begin{figure}[ht]
\caption{Negated feature global importance for predicting output of malign}\label{negglobalBenign}
\centering
\includegraphics[width=.6\textwidth]{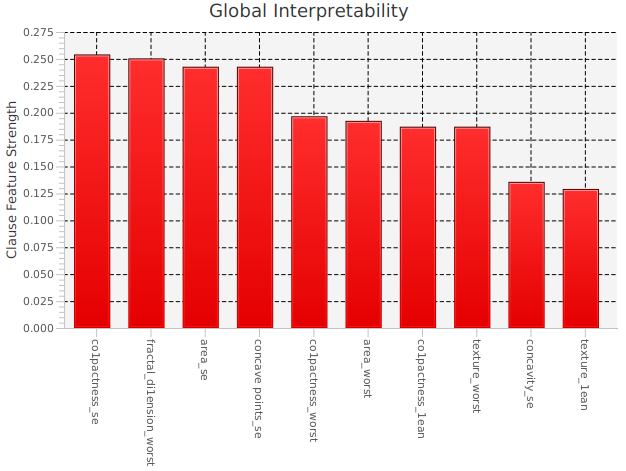}
\end{figure}

Finally, Figure \ref{negglobalBenign} reports the global importance for the features negated, calculated using $\bar{\phi}(f_u)$ in Eqn. \ref{global2} for the top features $f_u$. In essence, the strength of these features show that in general, the inclusion of their negated feature ranges had a positive impact on predicting the malign class of cells.

\subsection{Comparison with Neural Additive Models}

Recently introduced in \cite{agarwal2020neural}, Neural Additive Models offer a novel approach to general additive models. In all brevity, the goal is to learn a linear combination of simple neural networks that each model a single input feature. Being trained jointly, they learn arbitrarily
complex relationships between input features and outputs. In this section, we compare IWTM models to NAMs both in terms of AUC and interpretability. For convenience, we also show the performance against several other popular black-box type machine learning approaches.

We first investigate two classification tasks, continuing with a regression task to constrast interpretability against NAMs. All of the performance results, except for those associated with the IWTMs, are adapted directly from \cite{agarwal2020neural}. 

\subsubsection{Credit Fraud}

In this task, we wish to predict whether a certain credit card transaction is fraudulent or not. To learn such fraudulent transaction, we use the data set found in \cite{Pozzolo2015AdaptiveML} containing 284,807 transactions made by European credit cardholders. The data set is highly unbalanced, containing only 492 frauds (or 0.172 percent) of all transactions.

The training and testing strategy is to sample 60 percent of the true positives of fraudulent transactions, and then randomly select an equal number of non-fraudulent transactions from the remaining 284k+ transactions. Testing is then done on the remaining fraudulent transactions. 

\begin{figure}[ht]
\caption{Global feature importance in the credit card fraud detection data set. These are the 10 most influential features for predicting fraud from this data set.}\label{GlobalImportanceFraud}
\centering
\includegraphics[width=.6\textwidth]{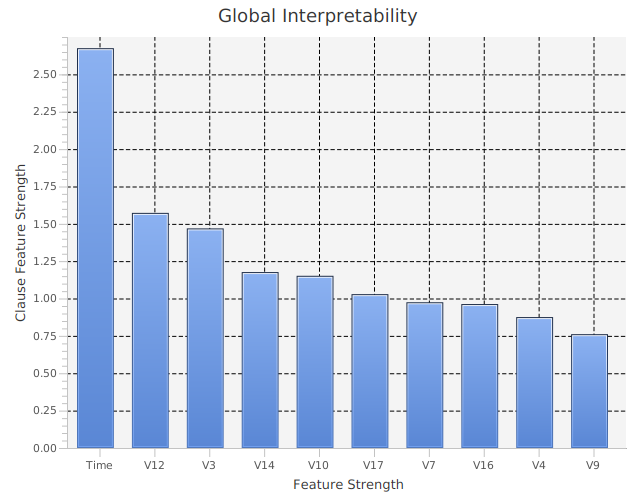}
\end{figure}

The globally important features for the fraudulent class of transactions are shown in Figure \ref{GlobalImportanceFraud}. With the exception of "time from first transaction" and "transaction amount", there is unfortunately no reference to other features names due to confidentiality. Thus, we cannot interpret features meaningfully. However, we can analyze how the ranges of the features vary according to predicting fraudulent or non-fraudulent transactions. 

Note that all plots concerning ranges have had the feature values normalized such that the minimal value for that feature is 0.0 and the maximal value is 1.0. We will use blue-colored bars to represent where the range with the most impact for the given class is located.

\begin{figure}[ht]
\caption{Feature ranges when predicting non fraudulent activity}\label{class0ranges}
\centering
\includegraphics[width=.6\textwidth]{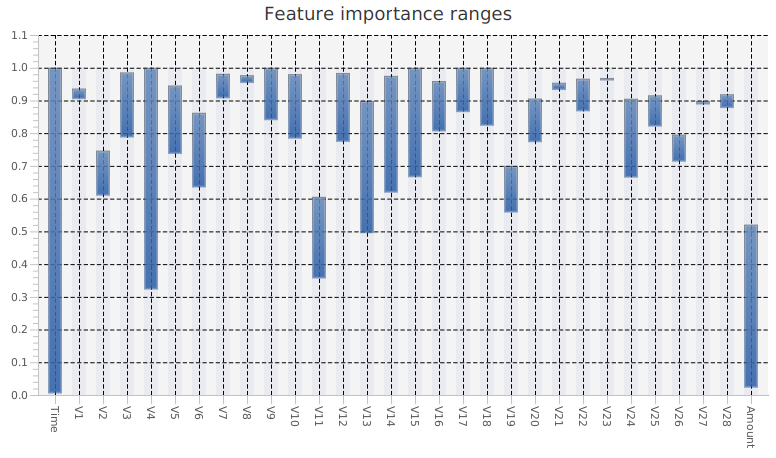}
\end{figure}

\begin{figure}[ht]
\caption{Feature ranges when predicting fraudulent activity. Notice that time is unbiased when predicting fraud}\label{class1ranges}
\centering
\includegraphics[width=.6\textwidth]{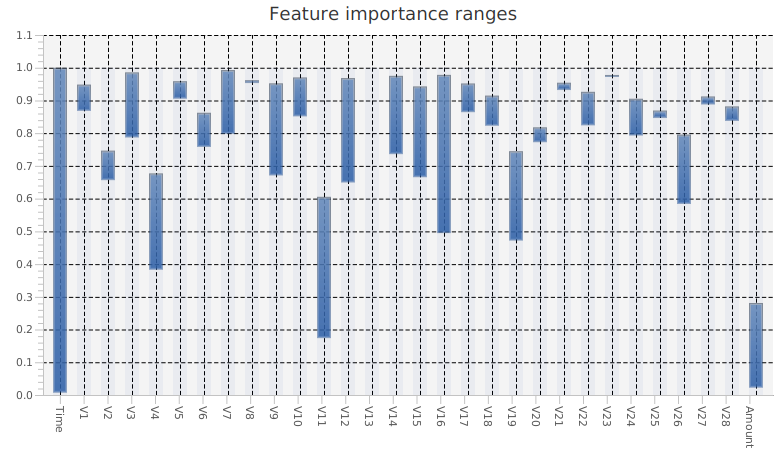}
\end{figure}

In applying the feature range expression from Eqn. \ref{global3} to a batch of non-fraudulent samples, we see from Figure~\ref{class0ranges} and Figure~\ref{class1ranges} that the "time from first transaction" feature, while one of the most important, is important in its entire range for both classes of transactions. Namely there is no range that sways the model to predict either fraudulent or non-fraudulent. This is in accordance with the NAM \cite{agarwal2020neural} (shown in Figure 12 as of the current draft on June 30 2020), where their graph depicts no sway of contribution into either class for the entire range of values.

The "transaction amount" feature can also be compared with NAMs, again considering  feature value ranges. In Figure~\ref{class0ranges}, we see that most relevant range is from around $0$ to $50$ for the non-fraudulent class, but for fraudulent transaction class shown in Figure \ref{class1ranges} the most salient range shrinks to about $0-25$. Indeed, for most of the other V-features, the value ranges tend to shrink, in particular for the higher value ranges. 

We can see that even though the visual representation of the model outputs in terms of feature and range importance are quite different, they can still be traced to yield similar conclusions.

\subsubsection{COMPAS: Risk Prediction in Criminal Justice}

\begin{figure}[ht]
\caption{Global feature importance with race removed as a feature when predicting low risk of re-offending}\label{compas_pred0}
\centering
\includegraphics[width=.6\textwidth]{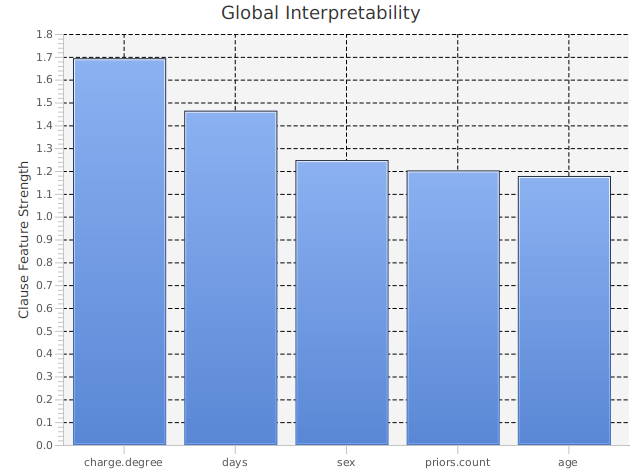}
\end{figure}

As a proprietary risk score developed to predict recidivism risk used to inform bail and sentencing and parole decisions, COMPAS is combined with a database containing the criminal history, prison time, demographics for defendants from Broward County from 2013 and 2014. In looking at the data features from ProPublica \cite{COMPAS}, it is clear that \verb|race| could be heavily biasing the predictions. Thus by removing \verb|race| as a feature, we are able to construct an interpretable model that demonstrated much more equality in importance across all remaining features. Figure \ref{compas_pred0} shows the global feature importance after race was removed as feature when predicting low probability of re-offending. Notice that the charge degree has the highest impact followed by number of days in jail, whereas age, sex and number of prior offenses have fairly close equal impact. 
In predicting higher probability of risk in re-offense, the number of prior offenses plays a more significant role, along with sex, while the charge degree is still the most important.  

\begin{figure}[ht]
\caption{Global feature importance with race removed as a feature when predicting high risk of re-offending}\label{compas_pred1}
\centering
\includegraphics[width=.6\textwidth]{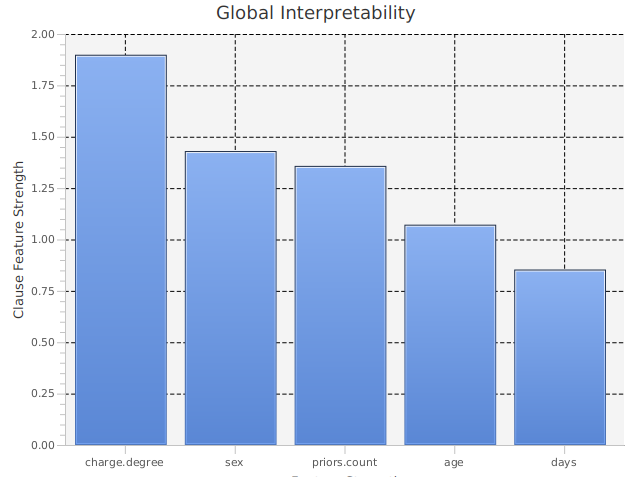}
\end{figure}

\begin{figure}[ht]
\caption{Ranges of features that have a high impact in predicting a high risk of re-offending}\label{compas_ranges}
\centering
\includegraphics[width=.6\textwidth]{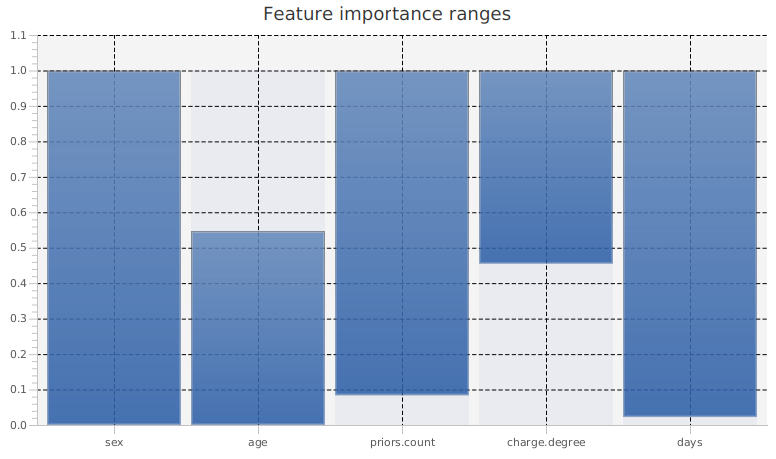}
\end{figure}

By looking at the value ranges that seem to influence high risk of re-offending, we get additional insight into the IWTM model (Figure \ref{compas_ranges}). We see that higher charge degree, higher prior convictions, and being male seems to give a higher risk. Number of days in jail/prison also contributes to a higher risk or re-offending for values above the bottom 5 percent level. 

It is worth noting that value of ranges of the prior convictions and the number of days in jail shown in Figure \ref{compas_ranges} seem to be in agreement with the ranges obtained with NAM \cite{agarwal2020neural}.

\subsubsection{Performance on Classification}

In terms of performance of IWTM in classification problems compared with six other approaches, we see that IWTMs have yielded competitive results.\footnote{The choice of hyperparameters of the IWTM can be summarized as picking the number of clauses randomly three times, between 50 and 500 clauses, with a threshold of twice the number of clauses. The best model in terms of accuracy was chosen of the three configurations.} In the COMPAS data, we prepared two separate models to observe performance. One of the models contains the \verb|race| feature bias in order to compare with results from other papers, and one with the bias removed. In Table \ref{classification_compare} we show how the two models compare with those used in \cite{agarwal2020neural} where \verb|race| was used a predictive feature in NAM and all other tested models therein. We see a slight dip in performance in the non-biased IWTM model (\verb|race|-feature removed), but it achieves competitive performance against all approaches when equipped with the same set of features.

\begin{table}
\caption{Performance comparison.}\label{classification_compare}
\centering
\begin{tabular}{|| c | c | c | c ||}
\hline
Method & COMPAS & Credit Fraud \\
\hline 
IWTM (w/out bias) & 0.739 (0.681) $\pm$ 0.018 (0.021) &  0.990 $\pm$ 0.005 \\
Logistic Regression & 0.730 $\pm$ 0.014 &  0.975 $\pm$ 0.010 \\
Decision Trees & 0.723 $\pm$ 0.010 &  0.956 $\pm$ 0.004 \\
NAMs & 0.741 $\pm$ 0.009  & 0.980 $\pm$ 0.002 \\
EBMs & 0.740 $\pm$ 0.012 & 0.976 $\pm$ 0.009 \\
XGBoost & 0.742 $\pm$ 0.009 & 0.981 $\pm$ 0.008 \\
DNNs & 0.735 $\pm$ 0.006 & 0.978 $\pm$ 0.003 \\
\hline
\end{tabular}
\end{table}

\subsubsection{California Housing Price data set}

We next investigate performance and interpretability on a regression task, using the California Housing data set. This data set appears in \cite{RePEc:eee:stapro:v:33:y:1997:i:3:p:291-297} where it is used to explore important features from a regression on housing price averages, offering features that are intrinsically interpretable and easy to validate in how they contribute to predictions. The data is comprised of one row per census block group, where a block group is the smallest geographical unit for which the U.S. Census Bureau publishes sample data, with a population typically between 600 to 3,000 people. The data set suggests we can derive and understand the influence of community characteristics on housing prices by predicting the median price of houses (in million dollars) in each California district.

We include results on both global and local interpretability, while also including results on influential ranges conditional on three tiers of prediction of housing prices (low, medium, high). 

The global feature importance diagram measures the importance of each feature by how often if appears as a positive contribution in the clauses, per Eqn. \ref{global2}. In summary, the diagram indicates that both median income and house location are the most influential features to predicting housing prices. This also conforms with the features derived by the NAM approach in \cite{agarwal2020neural}.

\begin{figure}[ht]
\caption{Global feature importance output by the Integer Weighted Tsetlin Machine}\label{globalImportanceHousing}
\centering
\includegraphics[width=.6\textwidth]{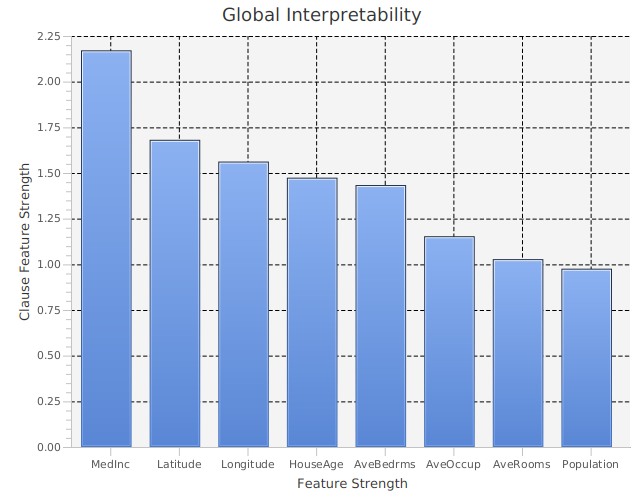}
\end{figure}

\begin{figure}[ht]
\caption{Example for local importance output: A neighborhood in an isolated block in the hills near Berkeley, with a housing price of 4.52 million. USD}\label{localImportanceHousing} 
\centering
\includegraphics[width=.6\textwidth]{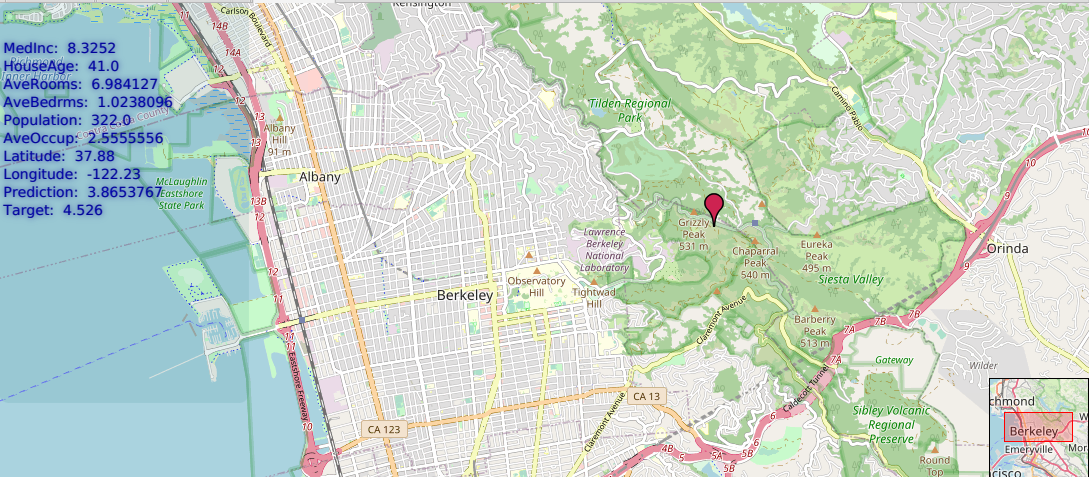}
\end{figure}

We now investigate local interpretability using a specific housing price instance. For our example, we take geographical region in the scenic hill area just east of Berkeley, known for its high real estate.  Being from an affluent Bay Area zip code, it should be clear that with a high housing price predicted, both median income and location should be the main drivers. Figure \ref{localImportanceHousing} shows the approximate area of the US Census block, with the exact input features given in the upper left of the figure.    

\begin{figure}[ht]
\caption{Output of the local feature importance showing median income and latitude (Bay Area) as the highest importance of such prediction}
\label{localImportanceHousing2}
\centering
\includegraphics[width=.6\textwidth]{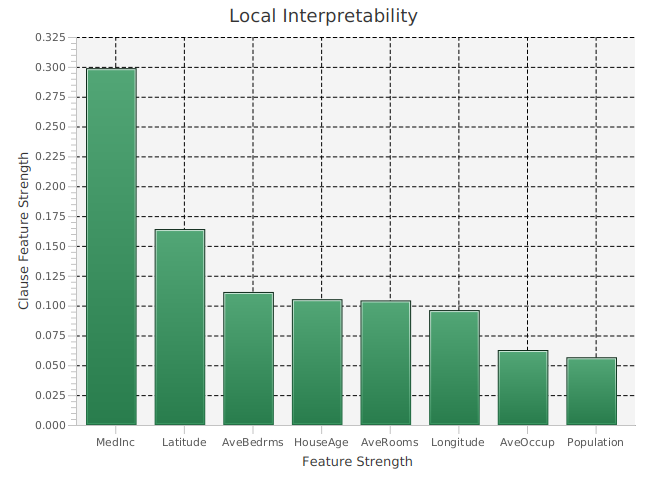}
\end{figure}

Applying the local interpretability expression from Eqn. \ref{locall4} to this specific sample, we see that the top features concur with the assumption on the particular real estate area (Figure \ref{localImportanceHousing2}). The location is sparsely populated with smaller occupancy per home, but hugely driven by the large median income and the Bay Area zip code. We also see that latitude achieves an exceptionally higher influence than longitude for this example, in comparison with their global importance, which could be from the fact that this particular latitude is aligned with northern California in the Bay Area and thus has more informative content to provide for a high housing cost prediction. Furthermore, housing age and number of bedrooms for this region in the Bay Area also seem to provide more information in making a prediction than longitude. This clearly exhibits how a local explanation can differ from a general global explanation of the model feature importance.  

\begin{figure}[ht]
\caption{Typical ranges which are influential when predicting lower-tiered housing prices.}\label{lowPrediction} 
\centering
\includegraphics[width=.6\textwidth]{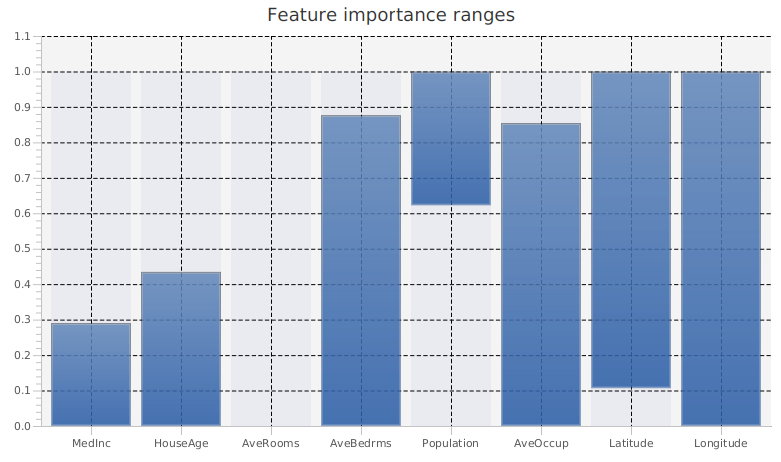}
\end{figure}

More insight can be gained by examining the feature value ranges that are important for certain levels of housing price predictions. Using the range importance expression Eqn. \ref{global3}, we look at some typical ranges of values that are attributed to lower prediction in the housing price. In Figure \ref{lowPrediction}, we note that Median income is low for this class of housing prices, while latitude and longitude values take on nearly the entire range of values. On the other hand, average number of rooms does not seem be a factor. 

\begin{figure}[ht]
\caption{Typical feature value ranges which are influential when predicting a mid-tiered housing prices.}\label{midPrediction} 
\centering
\includegraphics[width=.6\textwidth]{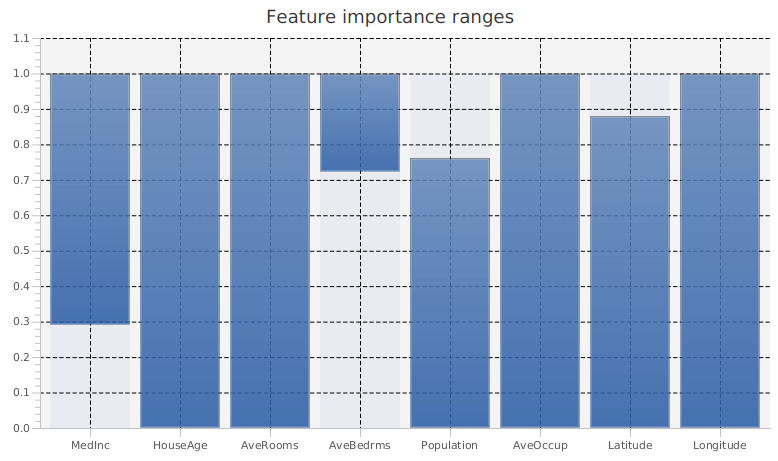}
\end{figure}

We now apply the same approach to mid-tiered housing prices (Figure \ref{midPrediction}). Here we expect to see higher median income and maybe some changes in the location to accommodate more rural housing prices. In general, relatively all ranges of values for most features are important, with the exception of median income which takes on the upper 70 percent of values. Further, the number of bedrooms plays a larger part in the prediction. Lower population seems to have a positive impact as well, accounting for those rural census blocks. 

Considering regression error, we compare techniques by means of average RMSE, obtaining the one-standard deviation figures via 5-fold cross validation. Table \ref{regression_performance} show the results compared with those from \cite{agarwal2020neural}, section 3.2.1.

\begin{table}
\centering
\caption{Comparison of performances of the regression IWTM with several other models}\label{regression_performance}
\begin{tabular}{|| c | c | c | c ||}
\hline
Method & California Housing  \\
\hline
IWTM & 0.572 $\pm$ 0.044  \\
Linear Regression & 0.728 $\pm$ 0.015  \\
Decision Trees & 0.720 $\pm$ 0.006 \\
NAMs & 0.562 $\pm$ 0.007  \\
EBMs & 0.557 $\pm$ 0.009  \\
XGBoost & 0.532 $\pm$ 0.014  \\
DNNs & 0.492 $\pm$ 0.009 \\
\hline
\end{tabular}
\end{table}

We see that IWTMs performs similarly to the other techniques, however, is outperformed by the DNNs. This can be explained by the low number of clauses employed, making it more difficult to exploit the large number of samples (20k), which gives the DNN \footnote{DNN with 10 hidden layers containing 100 units each with ReLU activation and Adam optimizer.} an advantage.

\subsection{Data dimension reduction and clustering}

In order to evaluate our dimension reduction and clustering scheme (Algorithm \ref{alg:reduc}), we apply the clustering procedure to a well known data set that has been used in other clustering approaches. It is the mice protein expression data from \cite{articleMice} consisting of the expression levels of 77 proteins/protein modifications that produced detectable signals in the nuclear fraction of cortex. The data includes experiments in the form of 15 measurements registered of each protein per sample/mouse that includes 38 control mice and 34 trisomic (Down syndrome) mice. For the control mice, there are thus 570 measurements, and for trisomic mice 510 measurements. The data set contains a total of 1080 measurements per protein with each measurement considered as an independent sample/mouse.

Being given that there are 77 various dimensions (features) for each sample, any visualization method of the data should be able to capture some clear clustering of the data where each cluster contains samples that share something in common with its nearest neighbors. As most visualization tools explore data in 2 or 3 dimensions, we will use 3 dimensions with 2 of the dimensions as unit-less x/y coordinates, and the 3rd given by a transparency value (less transparent means further away in the 3rd dimension).      

We visually compare with other clustering methods, which include t-Distributed Stochastic Neighbor Embedding (tSNE), standard PCA, and a contrastive PCA approach (CPCA) introduced in \cite{Abid2017ContrastivePC}. While these methods are not based on supervised learning and thus cannot be discussed in any quantitative comparisons, we wish to exhibit nonetheless the visual essence of how all of the approaches are clustering the data.

\begin{figure}[ht]
\caption{t-Distributed Stochastic Neighbor Embedding of the mice protein data}\label{tsneMice}
\centering
\includegraphics[width=.6\textwidth]{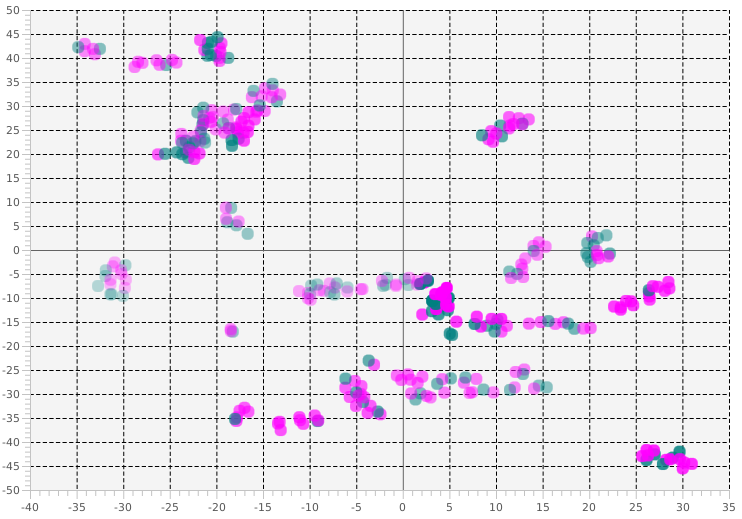}
\end{figure}

\begin{figure}[ht]
\centering
\includegraphics[width=.6\textwidth]{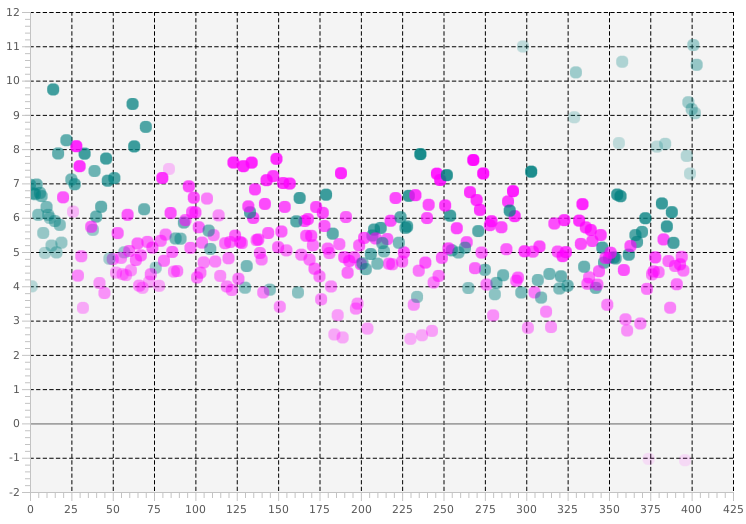}
\caption{Standard PCA transformation of mice protein data on 3 largest eigenvalue axis}\label{pcaMice}
\end{figure}

\begin{figure}[ht]
\centering
\includegraphics[width=.6\textwidth]{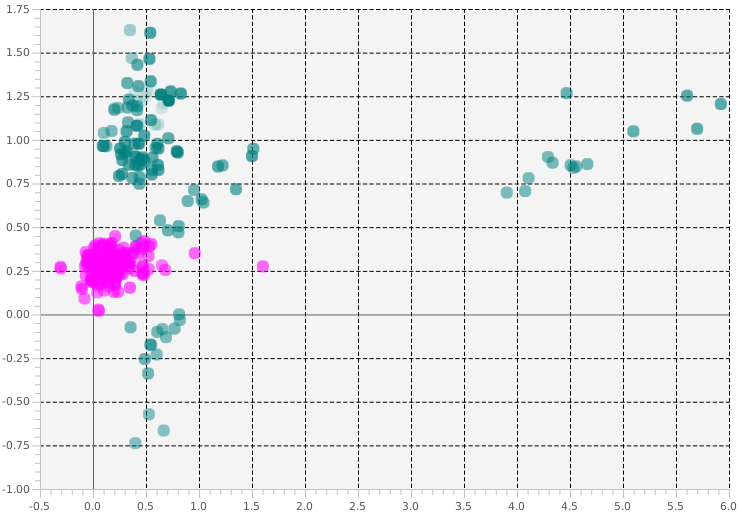}
\caption{Contrastive PCA transformation of mice protein data on 3 largest eigenvalue axis on contrasting data set.}
\label{fig:contrastiveMice}
\end{figure}

In the scatter plots in the Figures \ref{tsneMice}, \ref{pcaMice}, and \ref{fig:contrastiveMice}, the tSNE, PCA, and CPCA are all shown and plotted with the green values signifying healthy mice and the fuchia colored observations the trisomic mice. Clearly, the CPCA method accurately distinguishes between the two classes, while only using the protein markers as data. One cluster alone is attributed to all the trisomic mice. Using tSNE, while building clusters, it doesn't seem to distinguish between the two classes. Finally, PCA does not have any resemblance of distinguishable clustering at all. 

\begin{figure}[ht]
\centering
\includegraphics[width=.6\textwidth]{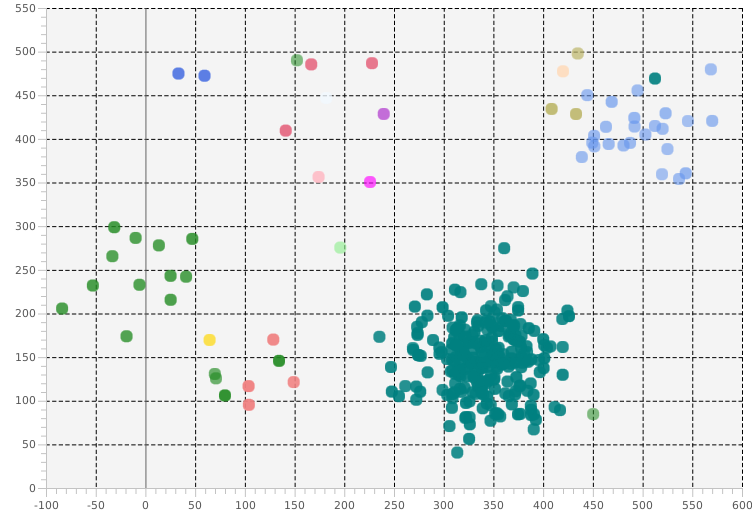}
\caption{Interpretable clustering and dimension reduction using Algorithm~\ref{alg:reduc}}\label{clusterMice}
\end{figure}

\begin{figure}[ht]
\centering
\includegraphics[width=.6\textwidth]{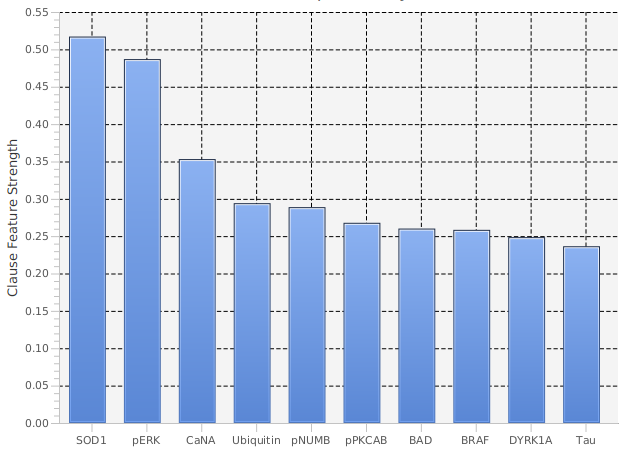}
\caption{Global interpretability of the mice protein expression for healthy mice}\label{globalImpMice}
\end{figure}

\begin{figure}[ht]
    \centering
    \subfloat[Observation chosen]{{\includegraphics[width=7cm]{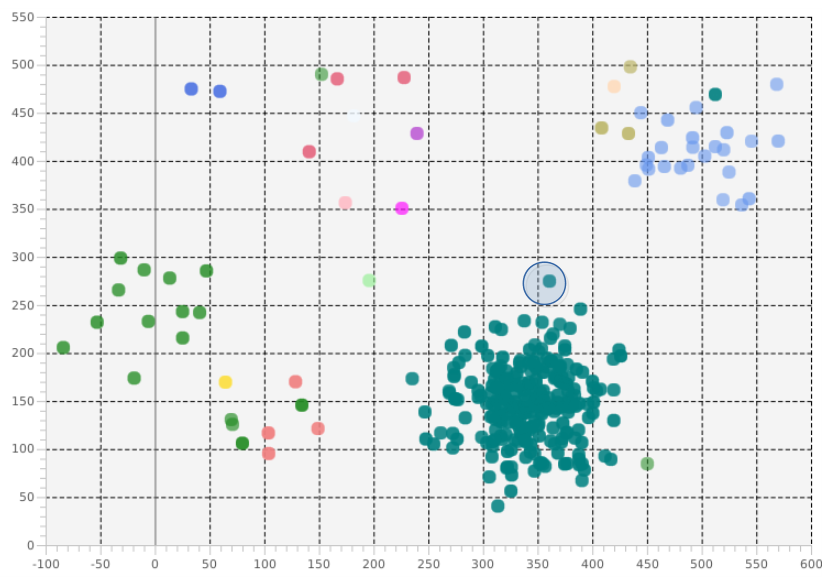} }}%
    \qquad
    \subfloat[Local interpretability]{{\includegraphics[width=7cm]{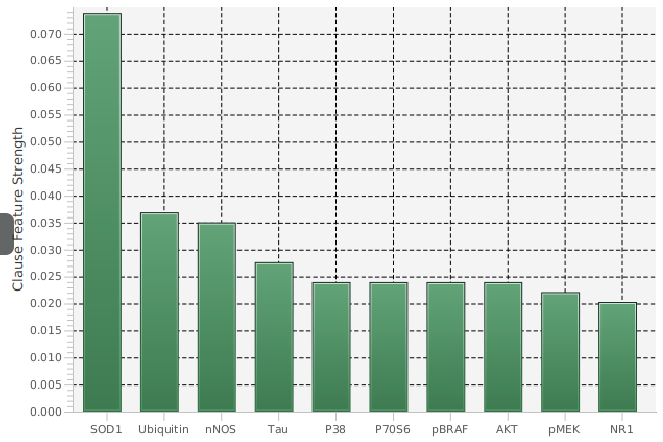} }}%
    \caption{Showing the local interpretability output on one particular observation in the largest cluster. All values in this cluster (teal colored) have SOD1 as the most impactful feature}%
    \label{fig:example}%
\end{figure}

\begin{figure}[ht]
    \centering
    \subfloat[Observation chosen]{{\includegraphics[width=7cm]{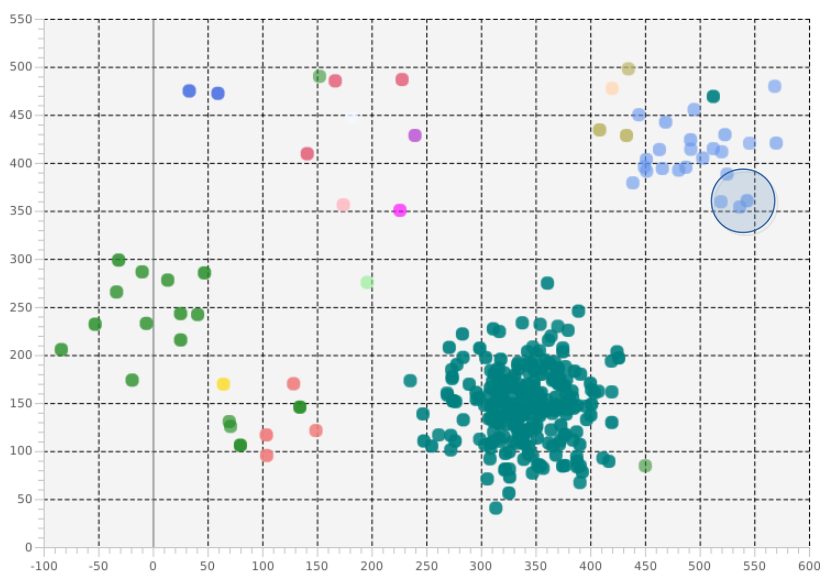} }}%
    \qquad
    \subfloat[Local interpretability]{{\includegraphics[width=7cm]{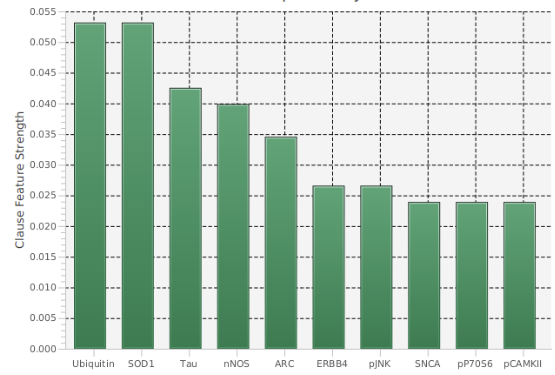} }}%
    \caption{Showing the local interpretability output on one particular observation in the largest cluster. All values in this cluster (green colored) have Ubiquitin as the most important feature}%
    \label{fig:example2}%
\end{figure}

\begin{figure}[ht]
    \centering
    \subfloat[Observation chosen]{{\includegraphics[width=7cm]{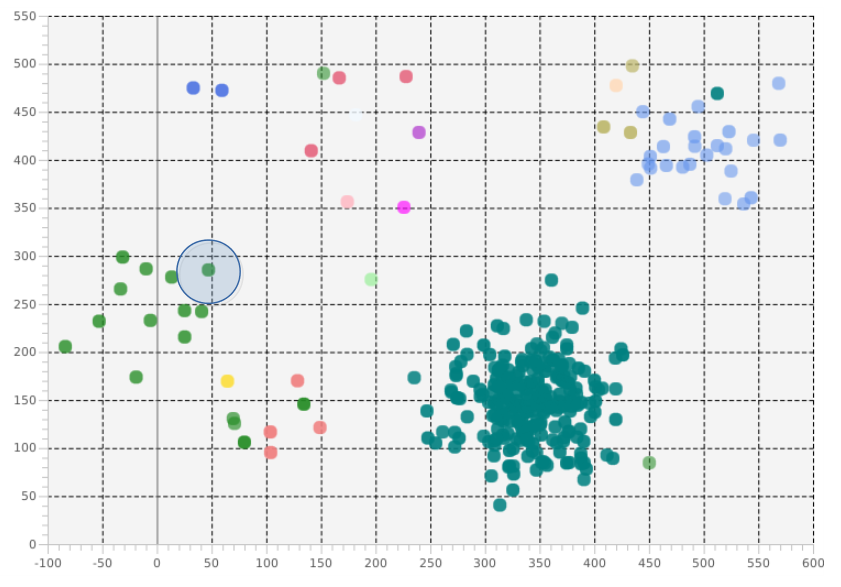} }}%
    \qquad
    \subfloat[Local interpretability]{{\includegraphics[width=7cm]{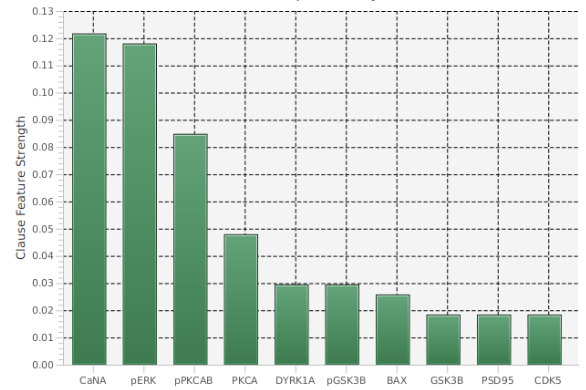} }}%
    \caption{Showing the local interpretability output on one particular observation in the largest cluster. All values in this cluster (light blue colored) have CaNA as the most important feature}%
    \label{fig:example3}%
\end{figure}

In the interpretable clustering approach, we visualized the results in the following manner. We first applied one epoch of TM learning. A weighted TM is applied to 70 percent of the data, where the label of either trisomic or healthy mouse was used to train the clauses. After the initial training, Algorithm \ref{alg:reduc} was applied, where we used the top 10 centroids in terms of their global importance coming from Eqn. \ref{global2}. Colors were then attributed to each of the 10 centroids, and any observation such that its local most influential feature had centroid $n$ as cluster center, the color of centroid $n$ was used.  Figure \ref{clusterMice} shows the resulting clustering scatter plot. Notice that a large portion of the clustering is centered around a very pertinent feature being SOD1, which happens to be globally the most important feature for predicting healthy mice, as shown in Figure \ref{globalImpMice}. Further, Figure \ref{fig:example} shows the local interpretability output on one particular observation in the largest cluster. All values in this cluster (teal colored) have SOD1 as the most impactful feature. Figure \ref{fig:example2} plots the local interpretability output on one particular observation in the largest cluster. All values in this cluster (light blue colored) have CaNA as the most important feature. Finally, Figure \ref{fig:example3}, depicts the local interpretability output on one particular observation in the largest cluster. All values inthis cluster (light blue colored) have CaNA as the most important feature.

\section{Conclusion}

In this paper, we proposed closed-form expressions for understanding why a TM model makes a specific prediction (local interpretability). Additionally, the expressions capture the most important features of the model overall (global interpretability). We further introduced expressions for measuring the importance of feature value ranges for continuous features. The expressions are formulated directly from the conjunctive clauses of the TM, making it possible to capture the role of features in real-time, also during the learning process as the model evolves.
Additionally, from the closed-form expressions, we derived a novel data clustering algorithm for visualizing high-dimensional data in three-dimensions. Finally, we compared our proposed approach against SHAP and state-of-the-art interpretable machine learning techniques. For both classification and regression, our evaluation show correspondence with SHAP as well as competitive prediction accuracy in comparison with XGBoost, Explainable Boosting Machines, and Neural Additive Models.

The clause structures that we investigated revealed interpretable insights about high-dimensional data, without the use of additional methods or additive approaches. Upon comparing the interpretability results to a staple in machine learning interpretability such as SHAP and a more recent novel interpretable extension to neural networks in NAMs, we observed that the TM-based interpretability yielded similar conclusions. Furthermore, we validated the approach of range importance on California housing data which offer human interpretable features. 

Current and future work on interpretability in Tsetlin Machines includes extending the interpretability expressions for local, global, and conditional range importance to also cover the convolutional TM approach \cite{granmo2019convolutional}. Further, a method for multivariate time-series analysis based on Tsetlin Machines \cite{blakelygranmoTS} needs to be developed. Finally, we are investigating a unsupervised approach to using Tsetlin Machines for clustering and outlier detection in real-time.  

\bibliographystyle{unsrt}  
\bibliography{references}

\begin{thebibliography}{10}

\bibitem{rudin2019stop}
Cynthia Rudin.
\newblock Stop explaining black box machine learning models for high stakes
  decisions and use interpretable models instead.
\newblock {\em Nature Machine Intelligence}, 1(5):206--215, 2019.

\bibitem{lundberg2017unified}
Scott Lundberg and Su-In Lee.
\newblock A unified approach to interpreting model predictions, 2017.

\bibitem{2016arXiv160204938T}
Marco {Tulio Ribeiro}, Sameer {Singh}, and Carlos {Guestrin}.
\newblock {``Why Should I Trust You?'': Explaining the Predictions of Any
  Classifier}.
\newblock {\em arXiv e-prints}, page arXiv:1602.04938, February 2016.

\bibitem{agarwal2020neural}
Rishabh Agarwal, Nicholas Frosst, Xuezhou Zhang, Rich Caruana, and Geoffrey~E.
  Hinton.
\newblock Neural additive models: Interpretable machine learning with neural
  nets.
\newblock 2020.

\bibitem{2019arXiv191102508S}
Dylan {Slack}, Sophie {Hilgard}, Emily {Jia}, Sameer {Singh}, and Himabindu
  {Lakkaraju}.
\newblock {Fooling LIME and SHAP: Adversarial Attacks on Post hoc Explanation
  Methods}.
\newblock {\em arXiv e-prints}, page arXiv:1911.02508, November 2019.

\bibitem{interpretabilitySteps2019}
Christian Blakely, Tayebeh Razmi, Bahar Sateli, and Christian Westermann.
\newblock {\em Five practical steps to make Artificial Intelligence (AI)
  interpretable}, 2019 (accessed July 27, 2020).

\bibitem{granmo2018tsetlin}
Ole-Christoffer Granmo.
\newblock {The Tsetlin Machine - A Game Theoretic Bandit Driven Approach to
  Optimal Pattern Recognition with Propositional Logic}.
\newblock 2018.

\bibitem{berge2019text}
Geir~Thore {Berge}, Ole-Christoffer {Granmo}, Tor~Oddbjørn {Tveit}, Morten
  {Goodwin}, Lei {Jiao}, and Bernt~Viggo {Matheussen}.
\newblock {Using the Tsetlin Machine to Learn Human-Interpretable Rules for
  High-Accuracy Text Categorization with Medical Applications}.
\newblock {\em IEEE Access}, 7:115134--115146, 2019.

\bibitem{abeyrathna2020regression}
K.~Darshana {Abeyrathna}, Ole-Christoffer {Granmo}, Xuan {Zhang}, Lei {Jiao},
  and Morten {Goodwin}.
\newblock {The Regression Tsetlin Machine - A Novel Approach to Interpretable
  Non-Linear Regression}.
\newblock {\em Philosophical Transactions of the Royal Society A}, 378, 2019.

\bibitem{granmo2019convolutional}
Ole-Christoffer Granmo, Sondre Glimsdal, Lei Jiao, Morten Goodwin, Christian~W.
  Omlin, and Geir~Thore Berge.
\newblock {The Convolutional Tsetlin Machine}.
\newblock 2019.

\bibitem{wheeldon2020learning}
Adrian {Wheeldon}, Rishad {Shafik}, Tousif {Rahman}, Jie {Lei}, Alex
  {Yakovlev}, and Ole-Christoffer {Granmo}.
\newblock {Learning Automata based Energy-efficient AI Hardware Design for
  IoT}.
\newblock {\em Philosophical Transactions of the Royal Society A}, 2020.

\bibitem{Narendra1989}
K~S Narendra and M~A~L Thathachar.
\newblock {\em {Learning Automata: An Introduction}}.
\newblock Prentice-Hall, Inc., 1989.

\bibitem{von2007theory}
John Von~Neumann and Oskar Morgenstern.
\newblock {\em Theory of games and economic behavior (commemorative edition)}.
\newblock Princeton university press, 2007.

\bibitem{Haugland2014}
Vegard Haugland, Marius Kj{\o}lleberg, Svein-Erik Larsen, and Ole-Christoffer
  Granmo.
\newblock {A two-armed bandit collective for hierarchical examplar based mining
  of frequent itemsets with applications to intrusion detection}.
\newblock {\em Transactions on Computational Collective Intelligence XIV},
  8615:1--19, 2014.

\bibitem{Granmo2007d}
Ole-Christoffer Granmo, B.~John Oommen, Svein~Arild Myrer, and Morten~Goodwin
  Olsen.
\newblock {Learning Automata-based Solutions to the Nonlinear Fractional
  Knapsack Problem with Applications to Optimal Resource Allocation}.
\newblock {\em IEEE Transactions on Systems, Man, and Cybernetics, Part B},
  37(1):166--175, 2007.

\bibitem{phoulady2020weighted}
Adrian Phoulady, Ole-Christoffer Granmo, Saeed~Rahimi Gorji, and Hady~Ahmady
  Phoulady.
\newblock {The Weighted Tsetlin Machine: Compressed Representations with
  Weighted Clauses}, 2019.

\bibitem{gorji2020indexing}
Saeed {Gorji}, Ole~Christoffer {Granmo}, Sondre {Glimsdal}, Jonathan {Edwards},
  and Morten {Goodwin}.
\newblock {Increasing the Inference and Learning Speed of Tsetlin Machines with
  Clause Indexing}.
\newblock In {\em International Conference on Industrial, Engineering and Other
  Applications of Applied Intelligent Systems}. Springer, 2020.

\bibitem{gorji2019multigranular}
Saeed~Rahimi {Gorji}, Ole-Christoffer {Granmo}, Adrian {Phoulady}, and Morten
  {Goodwin}.
\newblock {A Tsetlin Machine with Multigranular Clauses}.
\newblock In {\em Lecture Notes in Computer Science: Proceedings of the
  Thirty-ninth International Conference on Innovative Techniques and
  Applications of Artificial Intelligence (SGAI-2019)}, volume 11927. Springer,
  2019.

\bibitem{abeyrathna2019scheme}
K.~Darshana Abeyrathna, Ole-Christoffer Granmo, Xuan Zhang, and Morten Goodwin.
\newblock {A Scheme for Continuous Input to the Tsetlin Machine with
  Applications to Forecasting Disease Outbreaks}, 2019.

\bibitem{abeyrathna2020extending}
K.~Darshana Abeyrathna, Ole-Christoffer Granmo, and Morten Goodwin.
\newblock {Extending the Tsetlin Machine With Integer-Weighted Clauses for
  Increased Interpretability}.
\newblock 2020.

\bibitem{oommen1997stochastic}
B~John Oommen.
\newblock Stochastic searching on the line and its applications to parameter
  learning in nonlinear optimization.
\newblock {\em IEEE Transactions on Systems, Man, and Cybernetics, Part B
  (Cybernetics)}, 27(4):733--739, 1997.

\bibitem{shafik2020explainability}
Rishad {Shafik}, Adrian {Wheeldon}, and Alex {Yakovlev}.
\newblock {Explainability and Dependability Analysis of Learning Automata based
  AI Hardware}.
\newblock In {\em IEEE 26th International Symposium on On-Line Testing and
  Robust System Design (IOLTS)}. IEEE, 2020.

\bibitem{Tsetlin1961}
Michael~Lvovitch Tsetlin.
\newblock {On behaviour of finite automata in random medium}.
\newblock {\em Avtomat. i Telemekh}, 22(10):1345--1354, 1961.

\bibitem{Pozzolo2015AdaptiveML}
Andrea~Dal Pozzolo and Gianluca Bontempi.
\newblock Adaptive machine learning for credit card fraud detection.
\newblock 2015.

\bibitem{COMPAS}
ProPublica.
\newblock {\em COMPAS Analysis}, 2016 (accessed May 20, 2020).

\bibitem{RePEc:eee:stapro:v:33:y:1997:i:3:p:291-297}
Kelley Pace and Ronald Barry.
\newblock Sparse spatial autoregressions.
\newblock {\em Statistics and Probability Letters}, 33(3), 1997.

\bibitem{articleMice}
Clara Higuera, Katheleen Gardiner, and Krzysztof Cios.
\newblock Self-organizing feature maps identify proteins critical to learning
  in a mouse model of down syndrome.
\newblock {\em PloS one}, 10:e0129126, 06 2015.

\bibitem{Abid2017ContrastivePC}
Abubakar Abid, Vivek~Kumar Bagaria, Martin~Jinye Zhang, and James~Y. Zou.
\newblock Contrastive principal component analysis.
\newblock {\em ArXiv}, abs/1709.06716, 2017.

\bibitem{blakelygranmoTS}
Christian~D. Blakely and Ole-Christoffer Granmo.
\newblock {A Tsetlin Machine Approach for Learning and Prediction in
  Multivariate Time Series}.
\newblock 2020.

\end{thebibliography}

\end{document}